\documentclass[letterpaper, 10 pt, conference]{ieeeconf}  

\IEEEoverridecommandlockouts                              

\usepackage{epsfig} 
\usepackage[ruled,vlined,linesnumbered,noresetcount]{algorithm2e}
\usepackage{cite}
\usepackage{amsmath} 
\usepackage{amssymb}  
\usepackage{mathtools}
\usepackage{color}
\let\labelindent\relax
\usepackage{enumitem}
\usepackage{comment}

\usepackage{graphicx}
\usepackage[small]{caption}
\usepackage{subcaption}
\usepackage{url}
\usepackage{booktabs}

\title{\LARGE \bf
Stable bin packing of non-convex 3D objects with a robot manipulator}

\author{Fan Wang$^{1}$ and Kris Hauser$^{1}$
\thanks{$^*$This work is partially supported by an Amazon Research Award. }
\thanks{F. Wang and K. Hauser are with the Department of Electrical and Computer Engineering, Duke University,
        Durham, NC 27708, USA
        {\tt\small \{fan.wang2, kris.hauser\}@duke.edu}}%
}

\begin{document}

\maketitle
\thispagestyle{empty}
\pagestyle{empty}

\begin{abstract}

Recent progress in the field of robotic manipulation has generated interest in fully automatic object packing in warehouses. This paper proposes a formulation of the packing problem that is tailored to the automated warehousing domain. Besides minimizing waste space inside a container, the problem requires stability of the object pile during packing and the feasibility of the robot motion executing the placement plans.
To address this problem, a set of constraints are formulated, and a constructive packing pipeline is proposed to solve for these constraints. The pipeline is able to pack geometrically complex, non-convex objects with stability while satisfying robot constraints. In particular, a new 3D positioning heuristic called Heightmap-Minimization heuristic is proposed, and heightmaps are used to speed up the search. Experimental evaluation of the method is conducted with a realistic physical simulator on a dataset of scanned real-world items, demonstrating stable and high-quality packing plans compared with other 3D packing methods.

\end{abstract}
\section{Introduction}
\begin{figure}
\centering

    {\includegraphics[width = 3.9cm]{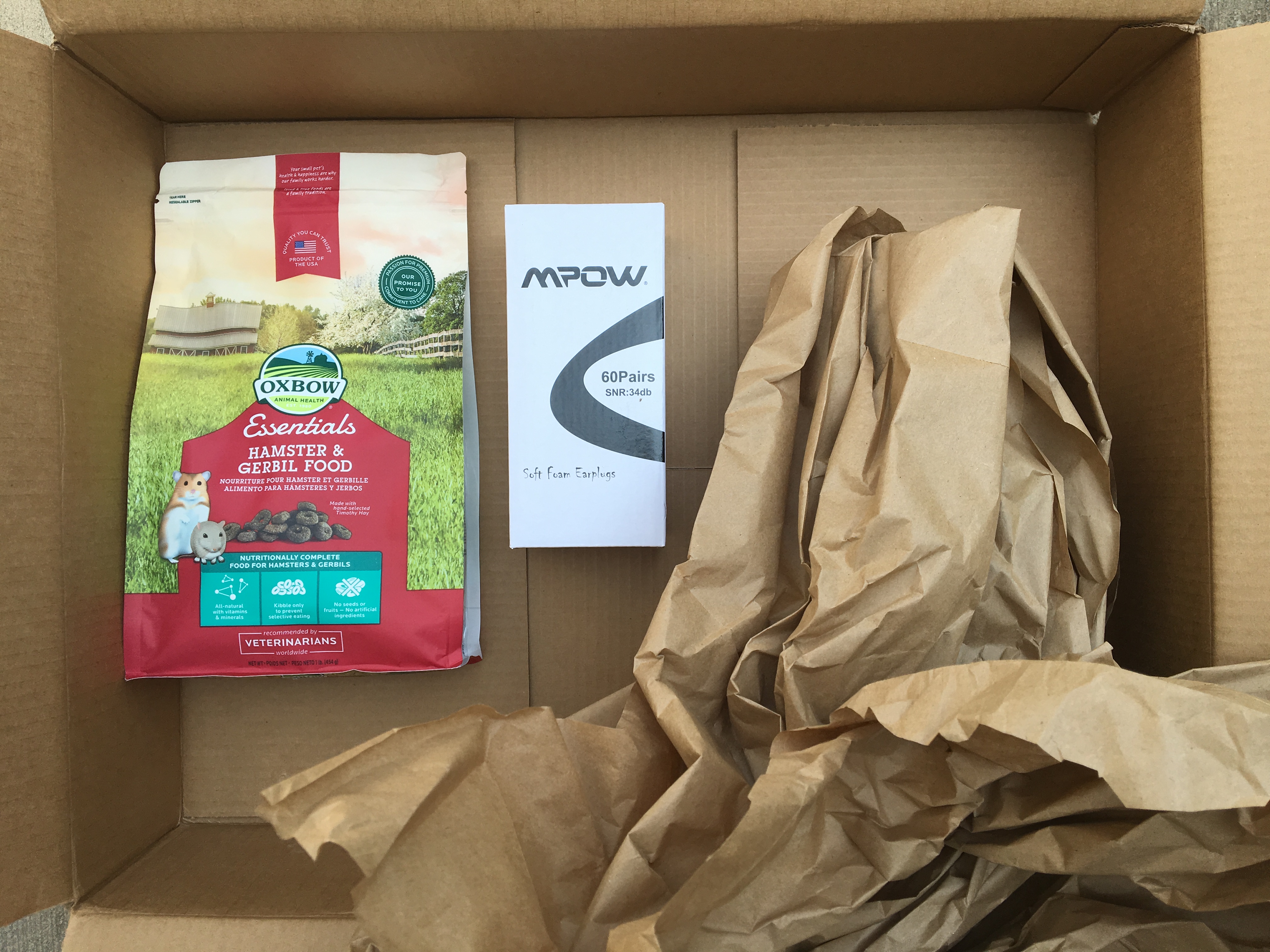}}
    {\includegraphics[width = 4cm]{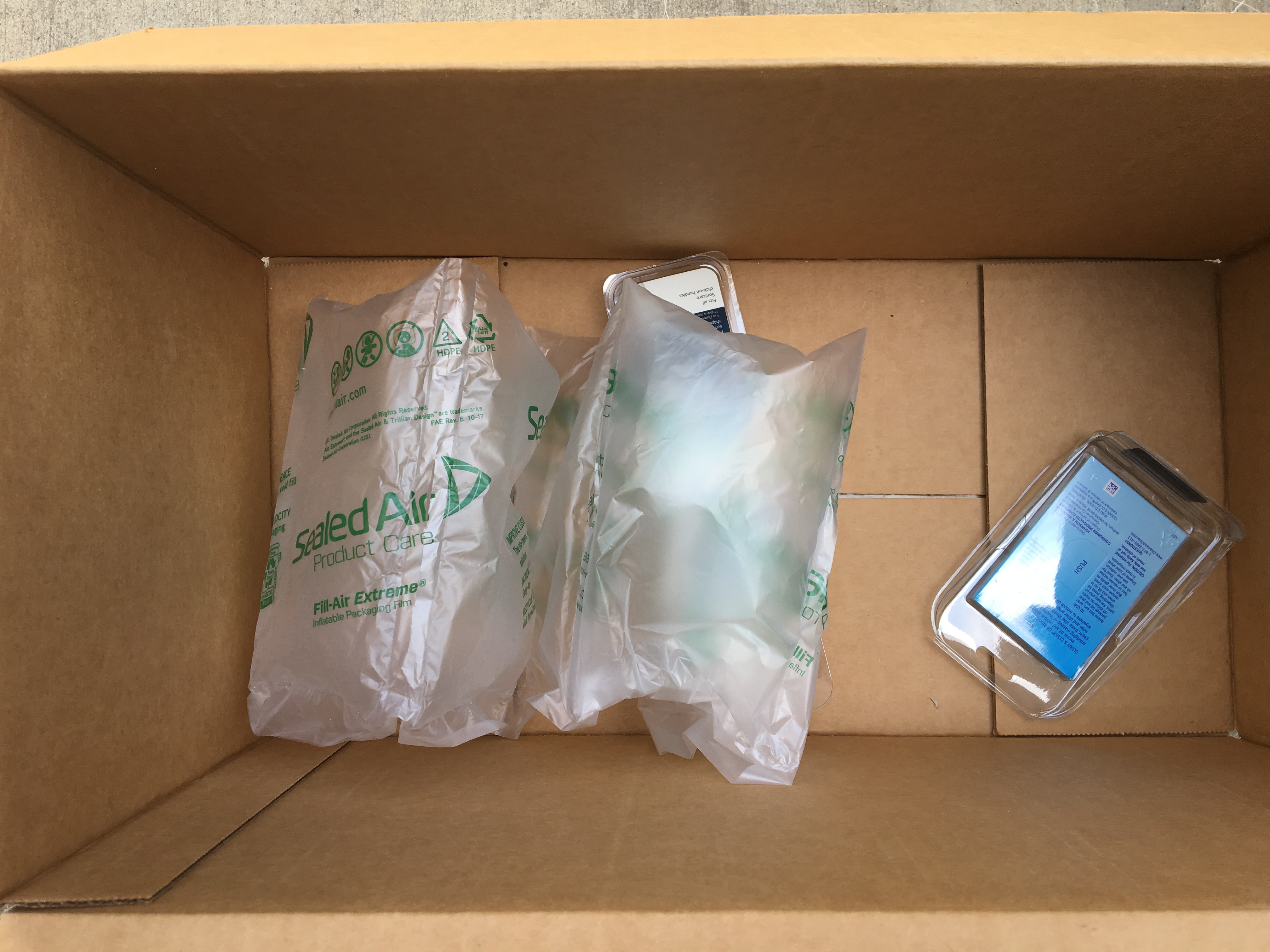}}

\caption{Examples of poor space utilization in shipping boxes.}
\label{fig:Amazonboxes}
\end{figure}

Recent years have seen increasing interest in warehouse automation, including fully automatic robot packing, supported by the technical progress made in the field of robotic manipulation, as demonstrated by recent competitions like the Amazon Robotics Challenge. The current state of practice in fulfillment centers leaves the responsibility of container selection and packing to human worker intuition. Due to demanding schedules, workers cannot employ much foresight in the packing process and are reluctant to re-pack. This commonly results in grossly oversized containers that generate waste and high shipping costs (Fig.~\ref{fig:Amazonboxes}). Better containers and packing plans could be chosen using automated algorithms, whether packing is accomplished by humans or robots.  

Problems that involve the placement of objects within a container or a set of containers are generally referred to as the cutting and packing problems. Most existing packing algorithms apply to idealized scenarios, such as rectilinear objects and floating objects not subject to the force of gravity. To perform automatic packing in warehouses using a pre-computed packing plan, several real-world issues need to be addressed, such as the feasibility of the packing under the force of gravity, and kinematics and clearance issues for the robot. If stability is not considered, the pile may shift during execution, and therefore subsequent placements are unlikely to be executed as planned. If kinematics and clearance are not considered, the robot may be asked to perform infeasible motions (e.g., grip an item from underneath, bring an item to a target through another interlocked item, or pass through the container wall).

For a packing plan to be feasible with a robot manipulator, a comprehensive set of constraints need to be formulated.

In addition to the two standard packing constraints:
\begin{enumerate}
    \item {\em Noninterference.} Each object is collision free,
    \item {\em Containment.} All objects are placed within the internal space of the container,
\end{enumerate}

we introduce the following constraints necessary for a robot-packable plan:

\begin{enumerate}
    \setcounter{enumi}{2}
    \item {\em Stability.} Each object is stable against previously packed objects and the bin itself, and
    \item {\em Manipulation feasibility.} A feasible robot motion exists to load the object into the target placement. The robot must obey kinematic constraints, grasp constraints, and collision constraints during this motion.
\end{enumerate}

In the following sections, we refer to constraints 1 and 2 as the {\em non-overlap} constraints, and constraints 1-4 as all constraints, or the {\em robot-packable} constraints.

While the application of {\em robot-packable} constraints is independent of the particular packing problem addressed, this paper focuses on the problem of offline packing of 3D irregular shapes into a single container, which is very relevant to an automated packing scenario in a fulfillment center, under {\em robot-packable} constraints.

To solve for the problem, we present the following main contributions:

\begin{enumerate}
    \item A polynomial time constructive algorithm to implement a resolution-complete search amongst feasible object placements, under {\em robot-packable constraints}.
    \item A 3D positioning heuristic named Heightmap-Minimization (HM) that minimizes the volume increase of the object pile as observed from the loading direction.
    \item A fast prioritized search scheme that first searches for {\em robot-packable} placement in a three-dimensional space that likely contains a solution, and falls back to search in a five-dimensional space.
\end{enumerate}

Our algorithm and others in comparison are tested in a realistic physics simulator, by packing large quantities of random item sets using high-quality, real-world object scannings. With item sizes of 3-5 objects (e.g., a common Amazon order size), the success rate is 100\% for finding and executing packing plans using small Amazon order boxes. Large number of items are also packed in stress tests, in these tests, 80\% of the placement plans were successfully executed in the physics simulator, which is significantly better than the 17\% success rate from a standard packing solver under the same testing condition. Empirical results also show that the new Heightmap-Minimization heuristic finds more placements than existing heuristics, both under {\em non-overlapping} constraints and {\em robot-packable} constraints.

\section{Related Work}
Popular variations of the cutting and packing problem include the bin and strip packing problem, the knapsack problem, the container loading problem, and others. Most existing research on cutting and packing handles floating 2D and 3D rectilinear objects under the {\em non-overlapping} constraints. Under some settings, such problems can be formulated and solved to optimally using the exact algorithms. One example of these state-of-the-art exact algorithms is the solution to the 3D bin packing problem using branch and bound, proposed by Martello et al.~\cite{vigo:2d,vigo:3d}, whose work is further extended by many including Boef et al.~\cite{den:robot} and Crainic et al.~\cite{teo:point}. The exact algorithms, although capable of finding the optimal solution if infinite time is spent, are strongly NP-hard~\cite{Garey1979ComputersAI} and do not guarantee optimal results within a reasonable amount of time, especially when a large number of instances are involved~\cite{vigo:3d}. Therefore, heuristic methods and metaheuristic approaches have been developed over the years, such as the popular “Bottom-Left (BL)” heuristic~\cite{baker:bl} and the Best-Fit-Decreasing heuristic~\cite{bestfit}.

On the other hand, irregular shape packing, often referred to as nesting, is a more recent variant of the cutting and packing problem. With non-rectilinear geometries, the search space is infinite, with few guidelines available to narrow it down to finite options. Metaheuristics such as Simulated Annealing (SA)~\cite{Kämpke1988, Zhang04, liu:hape} and Guided Local Search (GLS)~\cite{Faroe03, Egeblad:2009, Voudouris2003,Viegas15,BENNELL2013} are the most popular tools for solving a nesting problem. These methods commonly start with an initial placement and iteratively improve the placement by moving the pieces in the neighborhood while minimizing an objective function (e.g., overlap in the system). In addition to metaheuristic methods, recent work has also proposed constructive positioning heuristics for 3D irregular objects, such as Deepest-Bottom-Left-Fill (DBLF), which places items in the deepest, bottom-most, left-most position; and Maximum Touching Area (MTA), which places an item in a position that maximizes the total contact area of its faces with the faces of other items~\cite{wang:2h}.

Some research has taken additional constraints into consideration during packing. Egeblad et al., for example, use a two-stage GLS packing algorithm that, in the first stage, optimizes for the center of gravity and inertia of the pile and, in the second stage, minimizes overlap in the system. The optimization for different constraints is performed by adjusting the contribution of constraints dynamically in an augmented objective function~\cite{Egeblad:2009}; Liu et al. propose a constructive method called HAPE3D that packs irregular 3D shapes using a Minimum-Total-Potential-Energy heuristic~\cite{liu:hape}. This method performs a grid search in a 5D search space (e.g., $\phi$, $\psi$, $\theta$, X, Y) for the lowest gravitational center height z for each shape at the time of placement, and can be hybridized with metaheuristic SA to search for packing permutations that lead to lower total potential energy in the system. However, the proposed method is only a heuristic, and does not verify the stability of each placement. In contrast from these works, our method enforces stability explicitly using constraints.

We also know of one packing work that takes into account robot manipulation feasibility~\cite{den:robot}, in which the author proposes a variant of the orthogonal 3D box packing scheme such that no prior packed box is in front of, to the right of, or above the current placing box, to avoid possible collision with a vacuum gripper. Although this placing rule prevents a robot from colliding with boxes whose dimensions are much larger than the vacuum gripper, it cannot be generalized to other gripper geometries (e.g., parallel jaw gripper) and does not consider other aspects of robot feasibility such as kinematic constraints and graspability constraints. 

To the best or our knowledge, ours is the first packing work to solve for stability and robot-feasibility constraints simultaneously. Moreover, these constraints can be solved for arbitrary shaped 3D objects that are complex and non-convex.

\section{Problem Definition}

We address the problem of offline bin packing of 3D irregular shapes using a single box while ensuring the stability of each packed item and feasibility of the placement with a robot gripper.

Specifically, for a set $N$ geometries $\mathcal{G}_1,\dots,\mathcal{G_N}$ where $\mathcal{G}_i \subset \mathcal{R}^3$, let $\mathcal{C}$ donate the free space volume of the container and $\partial C$ as the boundary of the free space. Let $T_i\cdot \mathcal{G}_i$ denote the space occupied by item~$i$ when the geometry is transformed by $T_i$.  The problem is to find a placement sequence $S=(s_1,\ldots,s_N)$ of $\big\{1,\dots, N\big\}$ and transforms $\mathcal{T} = (T_1,\dots, T_N) $ such that each placement satisfies non-overlapping and containment constraints with geometries placed prior:
\begin{equation}
(T_i\cdot \mathcal{G}_i) \cap (T_j\cdot \mathcal{G}_j) = \emptyset, \; \forall i,j \in \big\{1,\dots, N\big\}, i\neq j
\end{equation}
\begin{equation}
T_i\cdot \mathcal{G}_i \subseteq \mathcal{C}, \; \forall i \in \big\{1,\dots, N\big\}
\end{equation}
and for each $k=1,\ldots,N$, stability constraints:
\begin{equation}
isStable\left(T_{s_k}\cdot \mathcal{G}_{s_k},  \mathcal{C}, T_{s_1}\cdot \mathcal{G}_{s_1}, \ldots, T_{s_{k-1}} \cdot \mathcal{G}_{s_{k-1}} \right)
\end{equation}
and manipulation feasibility constraints:
\begin{equation}
    isManipFeasible\left(T_{s_k} \cdot G_{s_k}, T_{s_1}\cdot \mathcal{G}_{s_1}, \ldots, T_{s_{k-1}} \cdot \mathcal{G}_{s_{k-1}} \right)
\end{equation}
It is important to note that both stability and manipulation feasibility constraints must be satisfied for {\em every intermediate arrangement} of objects, not just the final arrangement.

\subsection{Stability checking}

Stability is defined as the condition in which all placed items are in static equilibrium under gravity and frictional contact forces. We model the stack using point contacts with a Coulomb friction model with known coefficient of static friction.  Let the set of contact points be denoted as $c_1, \dots, c_K$, which have normals $n_1,\ldots,n_N$, and friction coefficients $\mu_1,\ldots,\mu_K$.  For each contact $c_k$, let the two bodies in contact be denoted $A_k$ and $B_k$.  Let $f_1,\ldots,f_K$ denote the contact forces, with the convention that $f_k$ is applied to $B_k$ and the negative is applied to $A_k$.  We also define $m_i$ as the mass of object $i$, and $cm_i$ as its COM.  We take the convention that either $A=0$ or $B=0$ indicates contact with the container wall.

The object pile is in static equilibrium if for $\forall i \in \big\{1,\dots, N\big\}$, there are a set of forces that satisfy the following conditions.

{\bf Force balance:}
 
\begin{equation}
    -\sum_{k\,|\,i=A_k} f_k + \sum_{k\,|\,i=B_k} f_k + m_i g = 0,
\end{equation}
 
{\bf Torque balance:}
\begin{equation}
    -\sum_{k\,|\,i=A_k} (cm_i - c_k) \times f_k +
    \sum_{k\,|\,i=B_k} -(cm_i - c_k) \times f_k = 0,
\end{equation}

{\bf Force validity:}
\begin{equation}
    f_k\cdot n_k > 0
\end{equation}
\begin{equation}\label{eq:frictioncone}
    \Vert f_k^\perp \Vert \leq \mu_k (f_k \cdot n_k)
\end{equation}
for all $k=1,\ldots,K$, where $f_k^\perp = f_k - n_k (f_k \cdot n_k)$ is the tangential component (i.e., frictional force) of $f_k$.

For a given arrangement of objects, an approximate set of contact points is obtained by slightly scaling geometries and determining the set of geometric features that overlap when scaled. We also perform a clustering step in which we merge all contact points within a grid size of 1\,cm, which reduces the number of contacts to a more manageable number.  Also, a pyramidal approximation for the friction cone is used, and the conditions above are formulated as a linear programming problem over $f_1,\ldots,f_N$, Our implementation uses the convex programming solver CVXPY to solve for a feasible set of forces~\cite{cvxpy}.  If no such forces can be found, the arrangement is considered unstable. 

\subsection{Manipulation feasibility}

This constraint checks feasibility of a packing pose when executed by a robot manipulator. This requires that the object be graspable from its initial pose and can be packed in the desired pose via a continuous motion, without colliding with environmental obstacles.

In our system we limit ourselves to existence of a feasible top-down placement trajectory within the grasp constraints. Robots performing pick and place (e.g., box packing) in the industry commonly use vertical motion that keeps the gripper opening parallel to the container~\cite{den:robot}. On this basis, we assume the robot gripper to perform top-down movement only when within the free space of the container. We also assume the existence of a grasp generator that produces some number of candidate end effector transforms, specified relative to an object's geometry that may be used to grasp the object. The pseudo-code for this procedure is given in Alg.~\ref{alg:ManipulationFeasibility}.

\begin{algorithm}
\SetKwInOut{Input}{input}
\SetKwInOut{Output}{output}
\SetAlgoLined
\setcounter{AlgoLine}{0}
\Input{Desired placed geometry $T \cdot \mathcal{G}$ and a set of grasp candidates $\{T_{1}^\mathcal{G},\dots T_{n}^\mathcal{G}\}$}
\Output{Whether a feasible top-down loading path exists}
 \For{$T^\mathcal{G} \in \{ T_{1}^\mathcal{G},\dots T_{n}^\mathcal{G} \}$}{
 
  Compute top-down end effector path $\mathcal{P}_{ee}$ interpolating from an elevated pose to a final pose $T\cdot T^\mathcal{G}$\;
 
  \For{$P_{ee} \in \mathcal{P}_{ee}$}{
   $f \gets $ IKSolvable ($P_{ee}$) $\wedge$ inJointLimits($P_{ee}$) $\wedge$ collisionFree($P_{ee}$)\;
   \If{$\neg f$}{Continue with Line 1}
  }
  \Return True
 }
 \Return False
 \caption{isManipFeasible}
 \label{alg:ManipulationFeasibility}
\end{algorithm}

\section{Pipeline for Robot-packable Planning}

We develop a constructive packing pipeline to solve for the set of {\em robot-packable} constraints proposed. Our algorithm accepts an item set, a container dimension, a constructive positioning heuristic, and\slash or a packing sequence, to produce packing plans. The pipeline packs each item to its optimized feasible pose in sequential order, without backtracking.

Our pipeline primarily consists of 4 components, namely:

\begin{enumerate}
    \item Placement sequence
    \item Positioning heuristic
    \item Stability check
    \item Manipulation feasibility check
\end{enumerate}

The pipeline implements a polynomial time resolution-complete search amongst feasible object placements under {\em robot-packable} constraints. Alternatively, the pipeline can produce results under a different level of constraints by enabling and disabling the stability and manipulation feasibility components that are implemented as separable processes.

The pipeline starts with a sequencing heuristic to sort all items in a tentative placement ordering and allocates them individually into the container in this sequence. For each object at the time of the allocation, a set of candidate transforms satisfying {\em non-overlap} constraints are generated given the container and object already placed. The candidate transforms are scored and ranked based on the positioning heuristic used. With no additional constraint required, the pipeline returns the best scored candidate transform. With additional constraints enabled, the ranked candidates go through specified constraint checks until a solution satisfying all required constraints is returned.
 
\subsection{Placement sequence}

Our algorithm allows user-specified packing sequences. If a sequence is not provided, we use a non-increasing bounding box volume heuristic to generate a tentative sequence, which is subject to adjustment if a solution cannot be found in its current ordering.

The non-increasing bounding box volume heuristic is equivalent to the non-increasing volume heuristic when applied on rectilinear objects. The latter is known to lead to the fastest convergence of the branch-and-bound algorithms~\cite{vigo:3d} and the good performance of the Best-Fit decreasing heuristics with rectangular-shaped objects. For 3D irregular shapes, bounding box volume is chosen instead of the exact volume to increase robustness against incomplete geometries and geometries that contain large concavity.


\subsection{Positioning Heuristic}

For a given item, a positioning heuristic (e.g., placement rule) identifies a free pose inside the container (or placement of the item) that is most preferred according to a specific criterion. Our pipeline accepts arbitrary positioning heuristics, but instead of applying the heuristic to obtain one optimal placement for each item, we use the score formulated from the positioning heuristic to rank candidate placements.

The candidate placements are obtained with a prioritized search among a discretized set of object poses. Instead of searching in the 6D space of SE(3), the algorithm first performs a grid search in a 3D space that likely contains stable solutions for horizontal surfaces.  This addresses the common case of packing on the first layer and on horizontal objects like boxes. The three-dimensional search space for a geometry $G$ is defined as follows: The rolls and pitches of $G$ are restricted to be a set of {\em planar-stable} orientations, which are a set of stable resting orientations of $G$ on a planar surface. We use the method of Goldberg et al.~\cite{goldberg:stable}, which uses the convex hull of the object, and also computes the likelihood of landing in a given orientation if the object is dropped onto a plane randomly. 

Our algorithm for finding a feasible placement of one object, given a set of rolls and pitches, is summarized in Alg ~\ref{alg:getSearchSpace}.
The height Z of the placement is analytically determined as the lowest legal placement for the oriented item at the given horizontal translation. A grid search is then performed for yaw, X and Y at a granularity that transforms $G$ within the container interior. If no robot-packable solution exists in this three-dimensional search space, the algorithm falls back to search in a 5D space where a grid search is performed for rolls and pitches as well. This fallback procedure is discussed in more detail in later sections.

\begin{figure}
\centering

    {\includegraphics[width = 8cm]{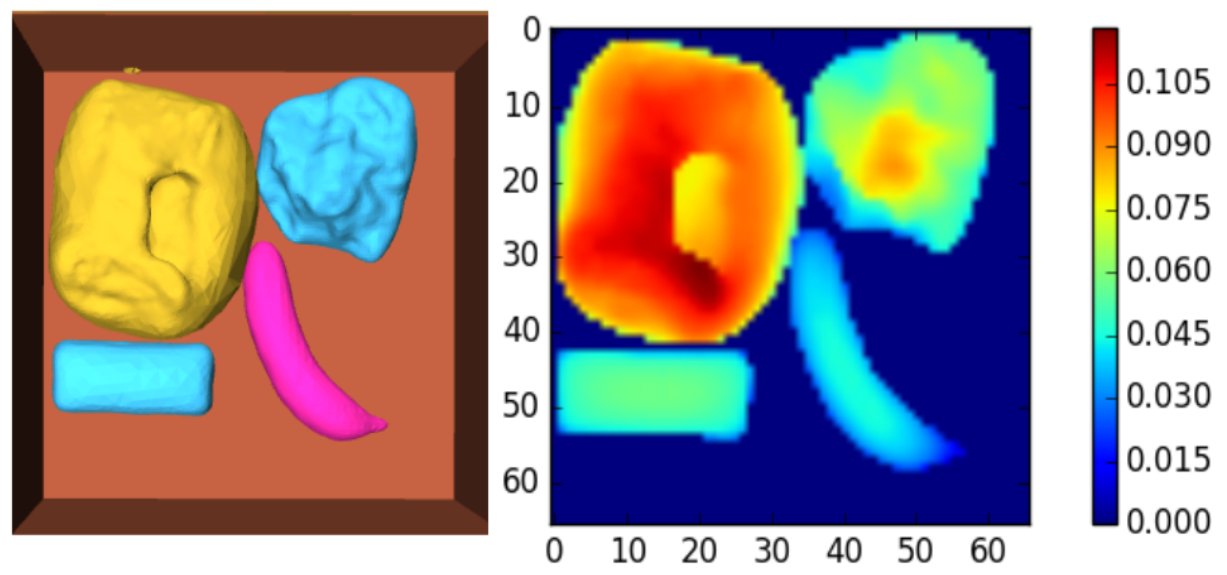}}
\caption{Terrain heightmap of 4 items.}
\label{fig:terrain}
\end{figure}

2D heightmaps are used to accelerate the computation of the lowest collision-free Z to an efficient 2D matrix manipulation.  Three heightmaps are computed: 1) a top-down heightmap $H_c$ of the container and objects already placed (Fig.~\ref{fig:terrain}), 2) a top-down heightmap $H_t$ of the object to be placed, measured relative to the lowest point at the object's given orientation, and 3) a bottom-up heightmap $H_b$ of the object to be placed, again measured relative to the lowest point. The container heightmap is obtained once at the beginning of object placement search, and an object heightmap is computed once for each distinct orientation in search.  The resolution of each heightmap is set equal to the resolution of search in the X, Y plane, and the object heightmaps are sized to the object's axis-aligned bounding box.  Raycasting is used to build these heightmaps, and rays that do not intersect with the object geometry are given height 0 in $H_t$ and $\infty$ in $H_b$.

Given an object orientation and X,Y location, we calculate the lowest collision-free Z as follows:
\begin{equation}
    Z = \max_{i=0}^{w-1} \max_{j=0}^{h-1} (H_c[x+i,y+j] - H_b[i,j])
\end{equation}
where $(x,y)$ are the pixel coordinates of $X,Y$, and $(w,h)$ to be the dimensions of $H_t$.

\begin{algorithm}
\SetKwInOut{Input}{input}
\SetKwInOut{Output}{output}
\SetAlgoLined
\setcounter{AlgoLine}{0}
\Input{Item geometry $\mathcal{G}$, container $\mathcal{C}$, rolls and pitches $O = \{(\phi_1,\psi_1),\dots,(\phi_n,\psi_n)\}$}
\Output{All legal candidate transforms $\mathcal{T} = \{(\phi_1,\psi_1,\theta_1,x_1,y_1,z_1),\dots,(\phi_n,\psi_n,\theta_n,x_n,y_n,z_n)\}$}
  \For{$(\phi,\psi) \in O$}{
   \For{$\theta \in \{0,\Delta r,2\Delta r,\ldots,2\pi-\Delta r\}$}{
    Let $R \gets R_z(\theta)R_y(\phi)R_x(\psi)$\;
    Discretize legal horizontal translations of $R \cdot \mathcal{G}$ into grid $\big\{(X_1,Y_1),\dots (X_n,Y_n)\big\}$\;
    \For{$(X,Y)$ in $\big\{(X_1,Y_1),\dots (X_n,Y_n)\big\}$}{
     Find the lowest collision free placement $Z$ at translation $X,Y$\;
     Let $T$ be rigid transform with rotation $R$ and translation $(X,Y,Z)$\;
     \If{$T \cdot \mathcal{G}$ lies within $\mathcal{C}$}{
      Add $T$ to $\mathcal{T}$
     }
    }
  }
  }
 \Return{$\mathcal{T}$}
 \caption{3DGridSearch}
 \label{alg:getSearchSpace}
\end{algorithm}

Once all legal candidate transforms are obtained, they are scored by a scoring function formulated from a positioning heuristic. For example, the Deepest-Bottom-Left-First heuristic can be formulated as the score:
\begin{equation}\label{eq:DBLF}
Z+ c \cdot (X+Y)
\end{equation}
where c is a small constant.

The candidates are then ranked by score (lower is better). If only {\em non-overlap} constraints are required, the placement candidate with the lowest score is returned. If additional constraints are specified, the ranked candidates will be checked for the additional constraints until a candidate satisfying all constraints is returned.

\label{sec:UpdateHeightmap}

After an object has been placed, we use a heightmap update subroutine that augments $H_c$ with the new object.  This subroutine is also used in our heightmap minimization heuristic. Given a pose $X,Y,Z$ of the object to be packed, and the top heightmap $H_t$ at the given orientation, we calculate an updated heightmap $H_c^\prime$ that contains the placed object as follows.  For all $i=0,\ldots,w-1$, $j=0,\ldots,h-1$, we let
\begin{equation}
    H_c^\prime[x+i,y+j] = \max(H_t[i,j] + Z,H_c[x+i,y+i]) 
\end{equation}
if $H_t[i,j]\neq 0$, and
\begin{equation}
    H_c^\prime[x+i,y+j] = H_c[x+i,y+i]
\end{equation}
otherwise.

\subsection{Pipeline summary and fall back procedures}

A packing attempt for a single item is summarized in Alg 3, where a set of rolls and pitches are given as input.  It sorts the candidate feasible placements $\mathcal{T}$ by the heuristic, and finds the first placement that satisfies stability and robot-packability.

The overall pipeline for packing multiple objects is given in Alg 4.  Given a heuristic packing sequence $S0$ (non-increasing bounding-box volume), it calls Alg. 3 for each item with the set of planar-stable rolls and pitches.  This first stage finds placements for most objects in typical cases.  For the remaining unpacked items $U$, the algorithm activates the {\em fallback procedure}.

The fallback procedure beginning in line 15 examines each unpacked item, and attempts to perturb the set of planar stable orientations to find an orientation that is packable.  It starts iterating over rolls and pitches until a solution is found, and if no solution is found the algorithm terminates with failure.  (Note that the first iteration begins by repacking the same item in the planar stable orientations, which may succeed now that the state of the bin has changed.)

\begin{algorithm}
\SetKwInOut{Input}{input}
\SetKwInOut{Output}{output}
\SetAlgoLined
\setcounter{AlgoLine}{0}
\Input{item geometry $\mathcal{G}$, container $\mathcal{C}$, pitches and yaws $O = \{(\phi_1,\psi_1),\dots,(\phi_n,\psi_n)\}$, sequence of the packed items $S=(s_1,\ldots,s_{i})$, transforms of the packed items $\mathcal{P} = \{P_1,\ldots,P_i\}$}
\Output{Transform $T$ or \textbf{None}}

$\mathcal{T} \gets$ 3DGridSearch($\mathcal{G},\mathcal{C},O$)\;
Score each $T$ in $\mathcal{T}$ based on heuristic used\;
\For{up to $N$ lowest values of $T$ in $\mathcal{T}$}{
  $s \gets $  isStable($T\cdot \mathcal{G}, \mathcal{C}, P_1\cdot \mathcal{G}_{s_1}, \ldots, P_i \cdot \mathcal{G}_{s_i}$)\;
  \If{$\neg s$}{{\bf continue}}
   Obtain grasp pose candidates $T_{1}^\mathcal{G},\dots T_{n}^\mathcal{G}$ compatible with $T$\;
   $f$ = isManipFeasible($T\cdot \mathcal{G}$, $(T_1^\mathcal{G},\dots T_{n}^\mathcal{G})$)\;
   \If{$f$} {\Return{$T$}}
  }
 \Return {\textbf{None}}
 
\label{alg:packOneItem}
\caption{packOneItem}
\end{algorithm}

\begin{algorithm}
\SetKwInOut{Input}{input}
\SetKwInOut{Output}{output}
\SetAlgoLined
\setcounter{AlgoLine}{0}
\Input{Item geometries $\mathcal{G}_1,\dots,\mathcal{G}_N$, container $C$, initial packing sequence $S0=(s0_1,\ldots,s0_N)$}
\Output{Transforms $\mathcal{T} = (T_1,\dots T_N)$ and final packing sequence $S=(s_1,\ldots,s_N)$, or \textbf{None}}
Initialize $\mathcal{T}, S, U,\mathcal{O} $ to empty lists\;

\For{$\mathcal{G}_i \in \{\mathcal{G}_1,\dots,\mathcal{G}_N\}$}{
    Get planar-stable rolls and pitches for $\mathcal{G}_i$ with the  top n highest quasi-static probabilities $O_i = \{(\phi_1,\psi_1),\dots,(\phi_n,\psi_n)\}$\;
     Add $O_i$ to $\mathcal{O}$\;
    }

\For{$s0_i \in \{s0_1,\dots, s0_N\}$}{
    $T = $ packOneItem($G_{s0_i}$, $C$, $O_{s0_i}$, $S$, $\mathcal{T}$)\;
    \uIf{$T$}{Add $T$ to $\mathcal{T}$\; Add $s0_i$ to $S$\;}
    \Else {Add $s0_i$ to $U$\;}
    }
\For{$u_i \in U$}{
    Let $\{(\phi_1,\psi_1),\dots,(\phi_n,\psi_n)\}$ be the planar-stable orientations in $O_{u_i}$\;
    
    \For{$t_r \in \{0, \Delta r,2\Delta r,\ldots,2\pi-\Delta r\}$}{\For{$t_p \in \{0,\Delta r,2\Delta r,\ldots,2\pi-\Delta r\}$}{
        $O^t = \{(\phi_1+t_r,\psi_1+t_p),\dots,(\phi_n+t_r,\psi_n+t_p)\}$\;
        $T = $ packOneItem($G_{u_i}$, $C$, $O^t$, $S$, $\mathcal{T}$)\;
        \If{$T$}{
            Add $T$ to $\mathcal{T}$\;
            Add $u_i$ to $S$\;
            {\bf continue} with Line 15}
        }
    }
    \Return ``no solution''
}
\Return ($\mathcal{T}, S$)
\label{alg:summary}
\caption{Robot-feasible packing with fall back procedures}
\end{algorithm}

\section{Heightmap-Minimization Heuristic}

The performance and solution quality of a multi-dimensional packing problem is highly susceptible to the item-positioning rule~\cite{Lodi2004}. However, existing positioning heuristics for 3D objects are scarce, and most of them are adapted directly from positioning rules developed for 2D packing problems, in which many were designed for packing rectangles only. It is known that naive generalization of heuristics from 2D to 3D leads to poor space utilization in the container \cite{LODI2002410,teo:point}.

Furthermore, multi-layer positioning rules for irregular shapes, particularly non-convex shapes, have not been sufficiently addressed in the previous literature. To address these shortcomings, we propose a novel positioning heuristic called the Heightmap-Minimization (HM) heuristic, which favors item placement that results in the smallest occupied volume in the container as observed from the loading direction.

Specifically, the score of a placement using HM heuristic is computed as follows.  Given the candidate transform $T = (roll, pitch, yaw, X, Y, Z)$, compute a tentative container heightmap $H_c^\prime$ using the update routine described in Sec.~\ref{sec:UpdateHeightmap}. Suppose its shape is $(w,h)$. The score for the placement using the HM heuristic is:
\begin{equation}\label{eq:scoring}
    c \cdot (X+Y) + \sum_{i = 0}^{w-1} \sum_{j = 0}^{h-1} H_c^\prime[i,j]
\end{equation}
where $c$ is a small constant.

\begin{figure}
\centering

    {\includegraphics[width = 7cm]{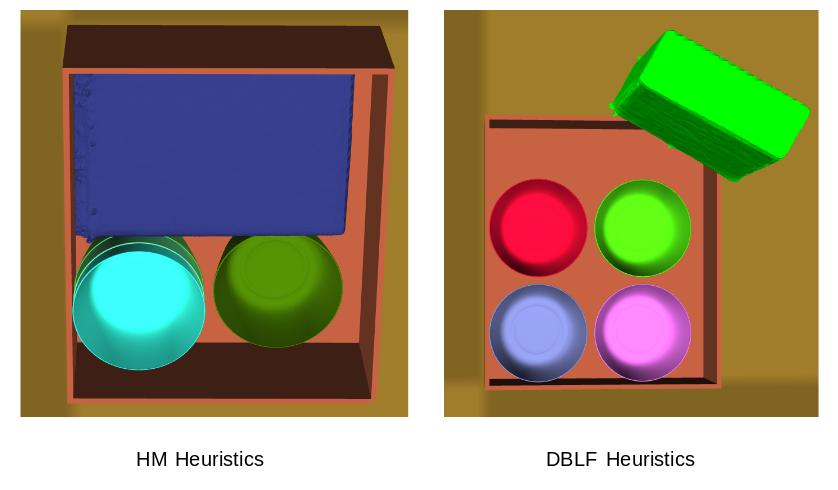}}

\caption{Packing a large box after three bowls using HM and DBLF heuristics. The HM stacks the bowls while the DBLF heuristics spreads them out, wasting space. Note that HM automatically selects the face-up orientation because it minimizes unused space, while the DBLF finds the same orientation to minimize center of gravity.}
\label{fig:heuristics}
\end{figure}

HM favors positions and orientations that result in good space utilization (Fig \ref{fig:heuristics}). Since hole fillings are expensive, and not allowed with heightmap representations, HM has the advantage of minimizing wasted space given the objects already placed in the container.

\begin{figure}
\centering

    {\includegraphics[width = 8cm]{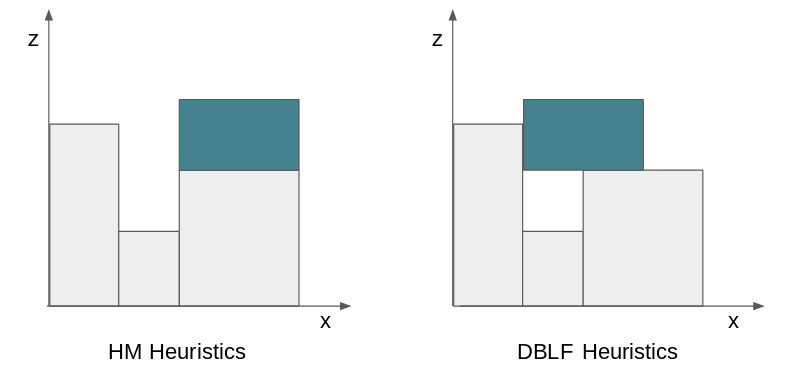}}

\caption{Most position-based heuristics ignore the wasted space underneath when an object is placed, such as DBLF. The HM heuristic, while maximizing unused space, inherently minimizes the wasted space between the terrain and the item to be placed and, therefore, likely increases contact areas and results in a more stable packing.}
\label{fig:heuristics_stability}
\end{figure}

HM also favors stable placements (Fig \ref{fig:heuristics_stability}) due to the maximum matching of geometry with the terrain underneath.  Because the bottom of the object is encouraged to match the shape of the supporting terrain, the generated placements are generally more stable than with other heuristics.

\section{Experiment}
We tested our algorithm using several distributions of item sets and heuristic choices. The resulting placements were then checked for feasibility in a physics simulator.

The 3D object models used are scannings of real-world objects drawn from the YCB object set\cite{YCB} and the APC 2015 object set\cite{APC}. 94 total items in the categories of toys, food, home supplies, etc. are used. The polygon meshes of each object on average contain 10,243 vertices.

Experiments are conducted on Amazon Web Services instance type m5.12xlarge. All computation times are measured on a single thread.
Parameters used in the experiment are: heuristic constant $c = 1$; heightmap resolution 0.002m; step size in both X and Y 0.01m; $\Delta r = pi/4$ in range $[0,\pi)$; friction coefficient $\mu = 0.7$. Contact points are obtained using the exact geometry with a scale factor of 1.03. The top 4 planar-stable rolls and pitches with the highest quasi-static probabilities are used. Up to 100 legal placement candidates are checked for stability and manipulation feasibility.

\subsection{Small Order Packing}

To evaluate our algorithm for everyday packing tasks seen in automated warehouses, we performed a small order packing test simulating problem settings in an Amazon warehouse. Per communication with personnel at Amazon, 3-5 items are a standard order size. We also selected the 5 container sizes used in the Amazon Robotics Challenge 2017~\cite{ARC}.

\begin{figure}
\centering

    {\includegraphics[width = 8.5cm]{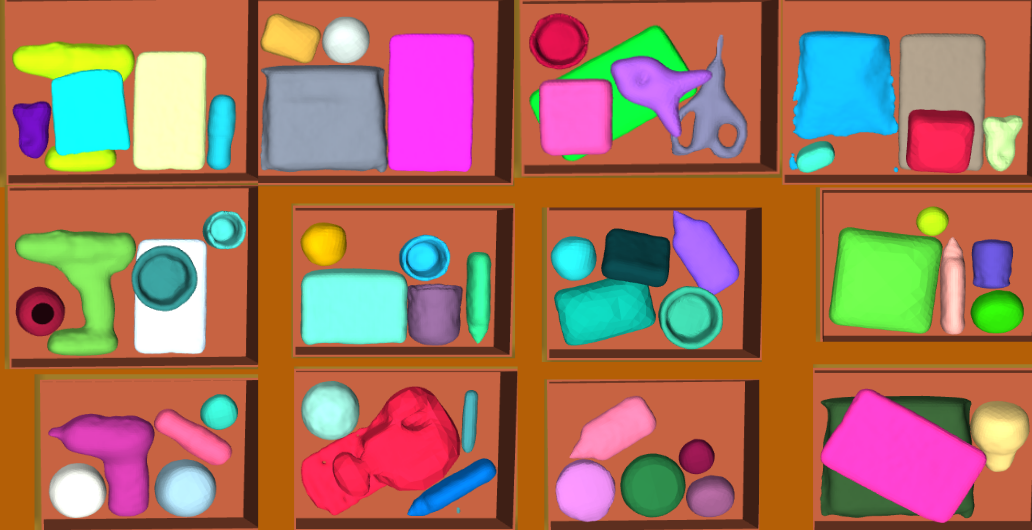}}
\caption{Examples of packing plans for item sets of size 3--5.}
\label{fig:Amazon}
\end{figure}

We generated 1000 item sets consisting of 3-5 models randomly selected from the APC and YCB datasets. The tested algorithm uses the HM heuristic and is required to obey all robot-packable constraints. The algorithm tries to find a feasible solution using the smallest container first, and if this step fails, the algorithm moves to the second smallest container and repeats the process until either a solution is found or all available containers are exhausted.
Using this process, solutions are found for 100\% of all orders using on average 8.54\,s per order (Fig.~\ref{fig:Amazon}).

\subsection{Comparisons on Large Item Sets}

Next, we compare how our method performs in more complex stress tests of item sets of size 10. A container of size $32\times 32 \times 30\,$cm is chosen, which is big enough to fit relatively large items while being small enough so that objects need to be packed in many layers to fit in the box. Item sets are randomly generated, and we attempt to find a non-overlapping placement for the set within the container using all of our tested methods.  (Robot-packability is not guaranteed.) This continues until 1000 feasible item sets are generated.

\begin{figure}
\centering
    \includegraphics[width = 8.5cm]{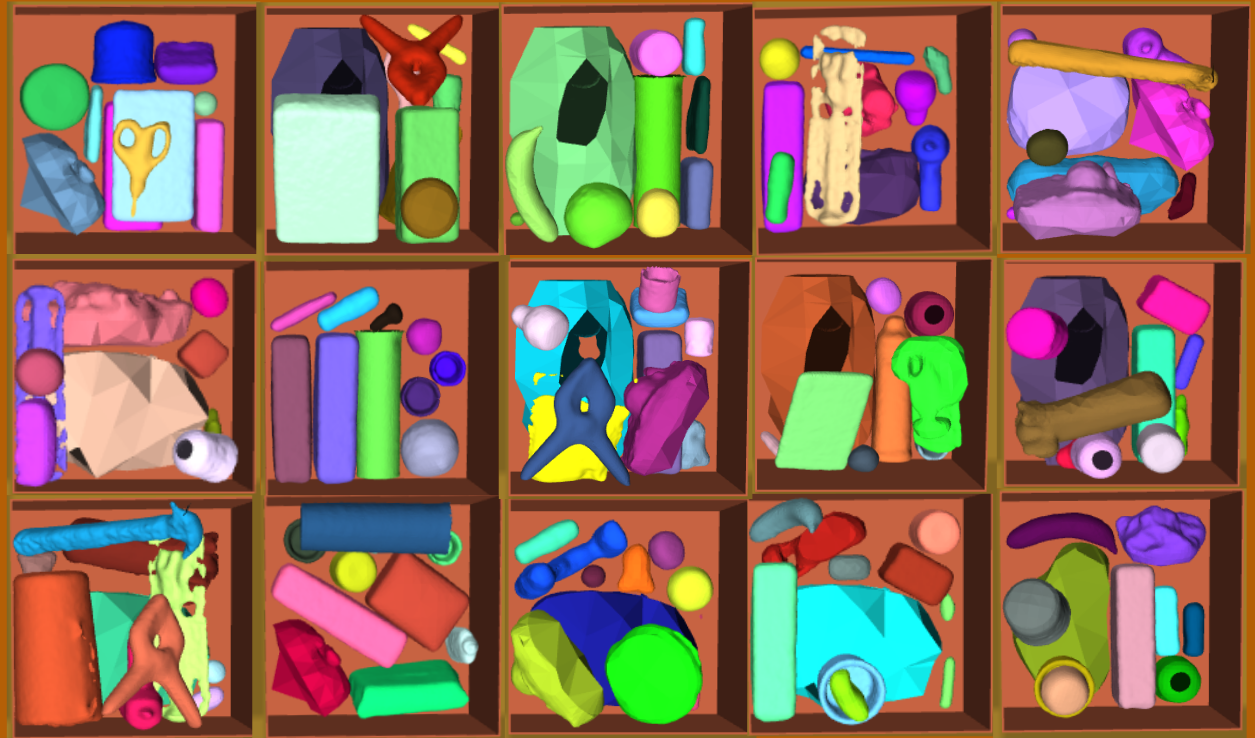}
\caption{Examples of packing plans for item sets of size 10.}
\label{fig:stablepacking10}
\end{figure}

We compare our HM heuristic against the DBLF and MTA heuristics~\cite{wang:2h}, as well as an implementation of a guided local search (GLS) method as described by Egeblad et al. \cite{Egeblad:2009}. The fast intersection area theorem, as described in Egeblad's paper, was not implemented. Therefore, for the fairness of the comparison, GLS was run with 5 random restarts, and each restart was terminated after 300\,s if a solution could not be obtained. Fig.~\ref{fig:stablepacking10} shows some example packing plans, and Table~\ref{table:Results} reports the percentage of solutions found and the average computation time for each set. With {\em non-overlap} constraints only, HM and DBLF have comparable high success rates and low running time, with HM finding solutions in 1.5\% more cases, while MTA and GLS are not as competitive. When {\em robot-packable} constraints are enabled, HM and DBLF still lead the percentage of solutions found (HM has a 1\% advantage over DBLF), both finding $> $96\% solutions, that is a 10\% edge over the third place MTA.  

The success rate for GLS under {\em robot-packable} constraints is not reported as such constraints were not implemented in the previous GLS works. Generally speaking, implementation of {\em robot-packable} constraints on GLS methods can be challenging due to two reasons: 1) Stability and feasibility constraint cannot easily be implemented in the objective function, and 2) GLS is not sequential so that there is no inherent packing sequence.

\setlength{\tabcolsep}{4pt}
\begin{table}[tbp]
\begin{center}
\caption{Comparing planning techniques on 10-item orders with and without robot-packability constraints}
\label{table:Results}
\begin{tabular}{@{}rccccc@{}}
\toprule
 & HM& DBLF~\cite{wang:2h} & MTA \cite{wang:2h} & GLS \cite{egeblad:2007}\\
\noalign{\smallskip}
\midrule
\noalign{\smallskip}
Success, non-overlap (\%) & 99.9 & 98.4 & 88.9 & 78.9  \\
Time, non-overlap (s) & 15.7 & 14.2 & 14.1 & 502\\
Success, all constraints (\%) & 97.1 & 96.3 & 86.3 & --- \\
Time, all constraints (s) & 34.9 & 50.1 & 95.4 & --- \\

\bottomrule
\end{tabular}
\end{center}
\end{table}
\setlength{\tabcolsep}{1.4pt}

Empirically, Heightmap-Minimization finds more solutions than any other method in comparison, under all degrees of constraint. With a {\em non-overlapping} constraint only, HM finds 99.9\% of all feasible solutions, leading the 2nd place DBLF heuristic by 1.5\%. After adding {\em manipulation feasibility} and {\em stability} constraints, each technique drops in success rate by a few percent, but HM still leads the 2nd place DBLF by 1\%.

With all constraints, HM has a mean average running time of 34.9\,s, which is at least 30\% shorter than other heuristics used. The minimum/maximum running time is 14.58/403.68\,s. This indicates that the highest ranked placements with HM heuristic are more likely to be stable placements than either of the other two heuristics.

In addition, only 3.2\% of the items are packed with the fallback procedure.  Of these, 51\% were packed by adjusting the packing sequence of the item, and the other 49\% were packed by performing the search in 5D. These statistics indicate that the 3D space searched is indeed highly likely to contain {\em robot-packable} solutions, and the prioritized search in 3D space is significantly more efficient than always searching in 5D. Our other tests indicate that simply performing the 5D search has an average running time of 522 \,s, which is 15 times slower than the prioritized search.

\subsection{Executing Complex Packing Plans }

Finally, we test open-loop execution feasibility of packing plans in the Klamp't robot physics simulator~\cite{klampt}. In the simulation, the robot places one item after another using a top-down loading direction. The plan is considered a success if: 1) when placing each object to its planned transform, the robot and the object do not collide with items placed prior, and 2) all items are contained within the free space of the container when placement is complete.

The robot used in the simulation is a Staubli TX90 robot, equipped with a rectangular vacuum gripper of 30\,cm length and 2\,cm diameter. We assume the vacuum gripper can grasp an object in its destination orientation at the center of the object's top surface, while the gripper axis is aligned with the Z axis. The robot executes top-down loading motions, ending in a pose where the item is elevated by 1cm from its planned transform; therefore there is an expected 1\,cm drop. We allow 20\,s for the items to settle before the next item is placed.

In the small item set case (3-5 items), 100\% of plans are successfully executed in the physics simulator, with an average drop of 1.08\,cm and 0.49\,cm horizontal displacement when executed in the physics simulator.

Table \ref{table:stability} describes results in the 10 item case. 768 out of 971 ($\approx$80\%) of {\em robot-packable} plans obtained with HM heuristic were executed successfully according to our success criteria. Using the non-overlap constraints only, the execution success rate was only 17\%.  
The shifts and drops of the item in the simulation are also logged. Under all constraints, the average drop of an item is 1.36 cm in the container (due to margins in the Z direction and extra 1cm lifted by the robot), and the horizontal shifts on average are 0.5 cm. Our method has a smaller displacement, which indicates greater stability, and a smaller drop, which reduces possible damage by falling from a height in real practice.

\setlength{\tabcolsep}{4pt}
\begin{table}
\begin{center}
\caption{Execution success rates in simulation, 10-item orders}
\label{table:stability}
\begin{tabular}{@{}rccc@{}}
\toprule\noalign{\smallskip}
 & Success (\%) & Drop (cm) & Horiz. Shift (cm) \\
\noalign{\smallskip}
\midrule
\noalign{\smallskip}
Non-overlap constraint& 17.11 & 1.95 & 1.29 \\
All constraint& 79.1 & 1.36 & 0.50 \\

\bottomrule
\end{tabular}
\end{center}
\end{table}
\setlength{\tabcolsep}{1.4pt}

\begin{figure}
\centering

    {\includegraphics[width = 8.5cm]{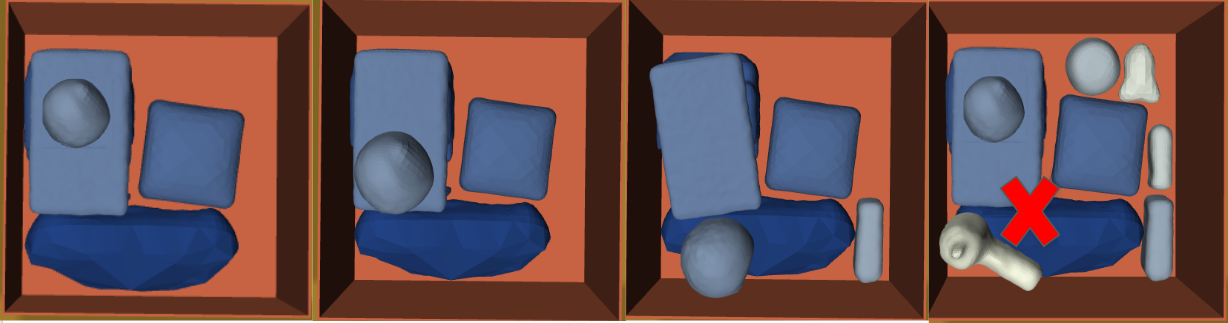}}
\caption{A typical execution failure case. The left three frames show a ball rolling out of its desired position. This prevents the subsequent placement of the drill, as shown in the rightmost frame.}
\label{fig:experiment8}
\end{figure}

The 20\% of failure cases are caused by an object falling out of its desired placement, which prevents subsequent items from being packed. The stability checker may be too optimistic, especially for intrinsically unstable objects like balls. Moreover, the impact of dropping an object could shift supporting objects.  An illustration is shown in Fig.~\ref{fig:experiment8}. 

\section{Conclusion}

In this paper, we address the automated packing problem in a warehouse setting in a well-constrained manner. A constructive pipeline is developed that can pack geometrically complex, non-convex objects with stability while satisfying robot constraints. A new Heightmap-Minimization heuristic is proposed as a positioning heuristic for efficient 3D irregular shape packing. Simulation results on exhaustive datasets demonstrate the effectiveness of the pipeline and the advantage of the new heuristic in finding stable and robot-packable plans. Robot-packable plans are shown to be far more successful in open-loop execution than the simple non-overlapping plans used in prior work.

Future work could address non-rigid objects or uncertainty in 3D scanned models.  Post-placement manipulation such as pushing could also increase the density of packing plans. We may also be able to increase the execution success rate by implementing a more conservative stability check, or to perform a closed-loop execution that replans once an object is observed to shift from its planned location.

\bibliographystyle{IEEEtran}
\bibliography{refs}

\begin{thebibliography}{10}
\providecommand{\url}[1]{#1}
\csname url@rmstyle\endcsname
\providecommand{\newblock}{\relax}
\providecommand{\bibinfo}[2]{#2}
\providecommand\BIBentrySTDinterwordspacing{\spaceskip=0pt\relax}
\providecommand\BIBentryALTinterwordstretchfactor{4}
\providecommand\BIBentryALTinterwordspacing{\spaceskip=\fontdimen2\font plus
\BIBentryALTinterwordstretchfactor\fontdimen3\font minus
  \fontdimen4\font\relax}
\providecommand\BIBforeignlanguage[2]{{%
\expandafter\ifx\csname l@#1\endcsname\relax
\typeout{** WARNING: IEEEtran.bst: No hyphenation pattern has been}%
\typeout{** loaded for the language `#1'. Using the pattern for}%
\typeout{** the default language instead.}%
\else
\language=\csname l@#1\endcsname
\fi
#2}}

\bibitem{vigo:2d}
S.~Martello and D.~Vigo, ``Exact solution of the two-dimensional finite bin
  packing problem,'' \emph{Management Science}, vol.~44, no.~3, pp. 388--399,
  1998.

\bibitem{vigo:3d}
S.~Martello, D.~Pisinger, and D.~Vigo, ``The three-dimensional bin packing
  problem,'' \emph{Operations Research}, vol.~48, no.~2, pp. 256--267, 2000.

\bibitem{den:robot}
E.~den Boef, J.~Korst, S.~Martello, D.~Pisinger, and D.~Vigo, ``Erratum to
  “the three-dimensional bin packing problem”: Robot-packable and
  orthogonal variants of packing problems,'' \emph{Operations Research},
  vol.~53, no.~4, pp. 735--736, 2005.

\bibitem{teo:point}
T.~G. Crainic, G.~Perboli, and R.~Tadei, ``Extreme point-based heuristics for
  three-dimensional bin packing,'' \emph{INFORMS Journal on Computing},
  vol.~20, no.~3, pp. 368--384, 2008.

\bibitem{Garey1979ComputersAI}
M.~R. Garey and D.~S. Johnson, ``Computers and intractability: A guide to the
  theory of np-completeness,'' 1979.

\bibitem{baker:bl}
B.~Baker, E.~Coffman, Jr., and R.~Rivest, ``Orthogonal packings in two
  dimensions,'' \emph{SIAM Journal on Computing}, vol.~9, no.~4, pp. 846--855,
  1980.

\bibitem{bestfit}
D.~Johnson, A.~Demers, J.~Ullman, M.~Garey, and R.~Graham, ``Worst-case
  performance bounds for simple one-dimensional packing algorithms,''
  \emph{SIAM Journal on Computing}, vol.~3, no.~4, pp. 299--325, 1974.

\bibitem{Kämpke1988}
T.~K{\"a}mpke, ``Simulated annealing: Use of a new tool in bin packing,''
  \emph{Annals of Operations Research}, vol.~16, no.~1, pp. 327--332, Dec 1988.

\bibitem{Zhang04}
D.~Zhang and W.~Huang, ``A simulated annealing algorithm for the circles
  packing problem,'' in \emph{Computational Science - ICCS 2004}, M.~Bubak,
  G.~D. van Albada, P.~M.~A. Sloot, and J.~Dongarra, Eds.\hskip 1em plus 0.5em
  minus 0.4em\relax Berlin, Heidelberg: Springer Berlin Heidelberg, 2004, pp.
  206--214.

\bibitem{liu:hape}
X.~Liu, J.-m. Liu, A.-x. Cao, and Z.-l. Yao, ``Hape3d---a new constructive
  algorithm for the 3d irregular packing problem,'' \emph{Frontiers of
  Information Technology {\&} Electronic Engineering}, vol.~16, no.~5, pp.
  380--390, May 2015.

\bibitem{Faroe03}
O.~Faroe, D.~Pisinger, and M.~Zachariasen, ``Guided local search for the
  three-dimensional bin-packing problem,'' \emph{INFORMS Journal on Computing},
  vol.~15, no.~3, pp. 267--283, 2003.

\bibitem{Egeblad:2009}
J.~Egeblad, ``Placement of two‐ and three‐dimensional irregular shapes for
  inertia moment and balance,'' vol.~16, pp. 789 -- 807, 06 2009.

\bibitem{Voudouris2003}
C.~Voudouris and E.~P.~K. Tsang, \emph{Guided Local Search}.\hskip 1em plus
  0.5em minus 0.4em\relax Boston, MA: Springer US, 2003, pp. 185--218.

\bibitem{Viegas15}
J.~L. Viegas, S.~M. Vieira, E.~M.~P. Henriques, and J.~M.~C. Sousa, ``A tabu
  search algorithm for the 3d bin packing problem in the steel industry,'' in
  \emph{CONTROLO'2014 -- Proceedings of the 11th Portuguese Conference on
  Automatic Control}, A.~P. Moreira, A.~Matos, and G.~Veiga, Eds.\hskip 1em
  plus 0.5em minus 0.4em\relax Cham: Springer International Publishing, 2015,
  pp. 355--364.

\bibitem{BENNELL2013}
J.~A. Bennell, L.~S. Lee, and C.~N. Potts, ``A genetic algorithm for
  two-dimensional bin packing with due dates,'' \emph{International Journal of
  Production Economics}, vol. 145, no.~2, pp. 547 -- 560, 2013.

\bibitem{wang:2h}
L.~Wang, S.~Guo, S.~Chen, W.~Zhu, and A.~Lim, ``Two natural heuristics for 3d
  packing with practical loading constraints,'' in \emph{PRICAI 2010: Trends in
  Artificial Intelligence}, B.-T. Zhang and M.~A. Orgun, Eds.\hskip 1em plus
  0.5em minus 0.4em\relax Berlin, Heidelberg: Springer Berlin Heidelberg, 2010,
  pp. 256--267.

\bibitem{cvxpy}
S.~Diamond and S.~Boyd, ``{CVXPY}: A {P}ython-embedded modeling language for
  convex optimization,'' \emph{Journal of Machine Learning Research}, vol.~17,
  no.~83, pp. 1--5, 2016.

\bibitem{goldberg:stable}
K.~Y. Goldberg, B.~Mirtich, Y.~Zhuang, J.~Craig, B.~Carlisle, and J.~F. Canny,
  ``Part pose statistics: estimators and experiments,'' \emph{IEEE Trans.
  Robotics and Automation}, vol.~15, pp. 849--857, 1999.

\bibitem{Lodi2004}
\BIBentryALTinterwordspacing
A.~Lodi, S.~Martello, and D.~Vigo, ``Tspack: A unified tabu search code for
  multi-dimensional bin packing problems,'' \emph{Annals of Operations
  Research}, vol. 131, no.~1, pp. 203--213, Oct 2004. [Online]. Available:
  \url{https://doi.org/10.1023/B:ANOR.0000039519.03572.08}
\BIBentrySTDinterwordspacing

\bibitem{LODI2002410}
\BIBentryALTinterwordspacing
------, ``Heuristic algorithms for the three-dimensional bin packing problem,''
  \emph{European Journal of Operational Research}, vol. 141, no.~2, pp. 410 --
  420, 2002. [Online]. Available:
  \url{http://www.sciencedirect.com/science/article/pii/S0377221702001340}
\BIBentrySTDinterwordspacing

\bibitem{YCB}
\BIBentryALTinterwordspacing
Ycb benchmarks – object and model set. [Online]. Available:
  \url{http://ycbbenchmarks.org}
\BIBentrySTDinterwordspacing

\bibitem{APC}
\BIBentryALTinterwordspacing
Rutgers apc rgb-d dataset. [Online]. Available:
  \url{http://pracsyslab.org/rutgers_apc_rgbd_dataset}
\BIBentrySTDinterwordspacing

\bibitem{ARC}
\BIBentryALTinterwordspacing
Amazon robotics challenge official rules. [Online]. Available:
  \url{https://www.amazonrobotics.com/site/binaries/content/assets/amazonrobotics/arc/2017-amazon-robotics-challenge-rules-v3.pdf}
\BIBentrySTDinterwordspacing

\bibitem{egeblad:2007}
J.~Egeblad, B.~K. Nielsen, and A.~Odgaard, ``Fast neighborhood search for two-
  and three-dimensional nesting problems,'' \emph{European Journal of
  Operational Research}, vol. 183, no.~3, pp. 1249 -- 1266, 2007.

\bibitem{klampt}
\BIBentryALTinterwordspacing
Klampt - intelligent motion laboratory at duke university. [Online]. Available:
  \url{http://motion.pratt.duke.edu/klampt}
\BIBentrySTDinterwordspacing

\end{thebibliography}

\end{document}